\documentclass[letterpaper, 10 pt, conference]{ieeeconf}
\IEEEoverridecommandlockouts
\usepackage{cite}
\usepackage{balance}
\usepackage{amsmath,amssymb,amsfonts}
\usepackage{graphicx}
\usepackage{textcomp}
\usepackage{xcolor}
\usepackage{booktabs}
\usepackage{caption}
\usepackage{subcaption}
\usepackage{algorithm}
\usepackage{algpseudocode}
\usepackage{tikz}
\usepackage{multirow}
\usepackage{transparent}
\usepackage{pgfplots}
\usepackage{fancyhdr}
\pgfplotsset{compat=1.18}
 
\def\BibTeX{{\rm B\kern-.05em{\sc i\kern-.025em b}\kern-.08em
    T\kern-.1667em\lower.7ex\hbox{E}\kern-.125emX}}
 
\begin{document}
 
\title{An Ensemble-Based Self-Taught Learning Approach for Parking Space Classification Under Limited Data}
 
\author{Lucas de Oliveira Cunha$^{1}$, Joelton Deonei Gotz$^{1}$, Paulo Lisboa de Almeida$^{2}$, Andre Gustavo Hochuli$^{3}$%
\thanks{$^{1}$Pontifícia Universidade Católica do Paraná, Computer Science, Curitiba, Paraná, Brazil
        {\tt\small \{c.oliveira25,joelton.gotz\}@pucpr.br}.}%
\thanks{$^{2}$Universidade Federal do Paraná, Department of Informatics,
        Curitiba, Paraná, Brazil
        {\tt\small paulorla@ufpr.br}.}%
\thanks{$^{3}$Pontifícia Universidade Católica do Paraná, Graduate Program in Informatics (PPGIa),
        Curitiba, Paraná, Brazil
        {\tt\small aghochuli@ppgia.pucpr.br}.}%
}
 
\maketitle
 \renewcommand{\headrulewidth}{0pt}
\renewcommand{\footrulewidth}{0pt}
\thispagestyle{fancy}
\fancyhf{}
\cfoot{\scriptsize Accepted for presentation at International Conference on Systems, Man, and Cybernetics (SMC 2026). \\ The final published version will be available on IEEE~Xplore.}
 
\begin{abstract}
Parking spot classification is a fundamental task in intelligent transportation systems, yet most deep learning approaches rely on large amounts of annotated data and exhibit limited generalization across heterogeneous environments. To address these limitations, we investigate a self-taught learning framework based on unsupervised representation learning with convolutional autoencoders. The proposed approach learns transferable visual representations from unlabeled data and reuses the learned encoders as fixed feature extractors for supervised classification with limited annotated samples in the target domain. To further enhance robustness and mitigate architectural bias, an ensemble of heterogeneous autoencoders is employed, with independent classifier heads and prediction fusion at inference time. Experiments conducted on the PKLot and CNRPark benchmarks under cross-dataset evaluation protocols show that the proposed ensemble-based strategy substantially reduces annotation requirements while improving robustness under significant domain shifts, achieving accuracies between 93\% and 96\% in data-constrained scenarios.
\end{abstract}
 
\IEEEpeerreviewmaketitle


\section{Introduction}\label{sec:intro}

The expansion of urban environments has increased the demand for intelligent transportation systems capable of efficiently monitoring parking infrastructure. Parking lot analysis is essential to this context, supporting tasks such as occupancy estimation and availability forecasting~\cite{bhattacharya2022review}. 

Recent advances in computer vision and deep learning have  improved performance in parking-related tasks~\cite{Almeida2022,Almeida2023Vehicle}. However, most existing approaches rely on large, scenario-specific annotated datasets and exhibit limited robustness when deployed across heterogeneous parking environments characterized by environmental variations and diverse scene layouts~\cite{amatoEtAl2017}, often resulting in significant performance degradation under unseen conditions~\cite{hochuli2023,Luza2025PKLOTDISTILLING}.

A major challenge in parking lot analysis is the high cost of manual annotation of data from the target parking lot\cite{hochuli2022} (e.g., collect and annotate as occupied or empty thousands of images from the target), often required by highly accurate systems \cite{Almeida2022, almeidaEtAl2015,amatoEtAl2017}. Such annotation is time-consuming and often impractical at scale, motivating learning paradigms that reduce reliance on labeled data while improving robustness to domain shifts.

In this vein, Self-Taught Learning (STL) has emerged as an effective paradigm to mitigate the reliance on large-scale annotated datasets\cite{Delazeri2025Representation, germani2023STL}. Unlike fully supervised methods, STL learns generic representations from unlabeled data and transfers them to target tasks with limited labeled samples.

In this work, we apply self-taught learning to parking spot classification and, as a contribution, introduce an ensemble of heterogeneous convolutional autoencoders to learn complementary unsupervised representations, improving robustness and generalization under cross-dataset parking scenarios.

To guide our investigation, we formulate the following research questions (RQs):
\begin{itemize}
\item (RQ1) To what extent can self-taught learning reduce the number of labeled target samples required to achieve competitive performance in parking lot classification tasks?
\item (RQ2) How robust are STL-based representations to domain shifts across different parking lots in terms of model generalization?
\item (RQ3) How does the proposed approach compare with fully supervised state-of-the-art methods?
\end{itemize}

To address these questions, we conduct a comprehensive experimental evaluation on the PKLot~\cite{almeidaEtAl2015} and CNRPark~\cite{amatoEtAl2017} benchmark datasets using a well-established cross-dataset evaluation protocol. The proposed self-taught learning framework is systematically assessed by varying the unlabeled source domain and the number of labeled target samples. This analysis evaluates robustness under domain shifts and demonstrates the feasibility of learning transferable parking representations with limited labeled target data.

\section{State of The Art}\label{sec:sota}

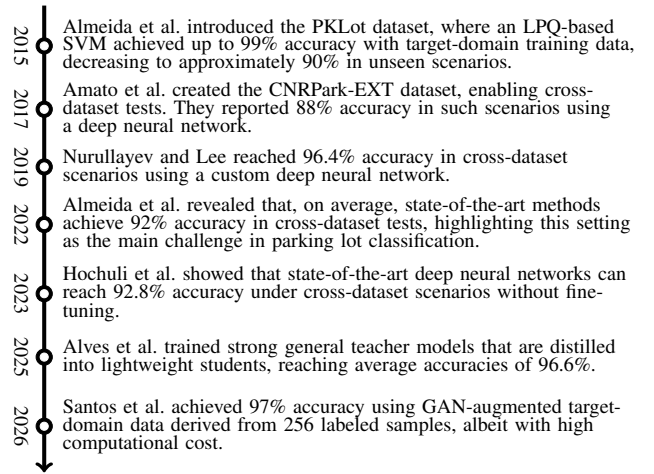
\begin{figure}[htb]  
   \centering  
    \resizebox{1\linewidth}{!}{\tikzset{
  textNode/.style={
  text width=10.0cm,
  align=left,
  anchor=west
  ,execute at begin node=\setlength{\baselineskip}{1pt}
  }
}

\begin{tikzpicture}
  \draw[->,black,line width=2] (0.9,8.7) -- (0.9,0.6);

  \filldraw[black] (0.9,8.0) circle (4pt);
  \filldraw[white] (0.9,8.0) circle (2pt);
  \node[rotate=-90] at (0.5,8.0) {2015};
  \node[textNode] at (1.1,8.0) {Almeida et al. introduced the PKLot dataset, where an LPQ-based SVM achieved up to 99\% accuracy with target-domain training data, decreasing to approximately 90\% in unseen scenarios.};

  \filldraw[black] (0.9,6.9) circle (4pt);
  \filldraw[white] (0.9,6.9) circle (2pt);
  \node[rotate=-90] at (0.5,6.9) {2017};
  \node[textNode] at (1.1,6.9) {Amato et al. created the CNRPark-EXT dataset, enabling cross-dataset tests. They reported 88\% accuracy in such scenarios using a deep neural network.};

  \filldraw[black] (0.9,5.9) circle (4pt);
  \filldraw[white] (0.9,5.9) circle (2pt);
  \node[rotate=-90] at (0.5,5.9) {2019};
  \node[textNode] at (1.1,5.9) {Nurullayev and Lee reached 96.4\% accuracy in cross-dataset scenarios using a custom deep neural network.};

  \filldraw[black] (0.9,4.9) circle (4pt);
  \filldraw[white] (0.9,4.9) circle (2pt);
  \node[rotate=-90] at (0.5,4.9) {2022};
  \node[textNode] at (1.1,4.9) {Almeida et al. revealed that, on average, state-of-the-art methods achieve 92\% accuracy in cross-dataset tests, highlighting this setting as the main challenge in parking lot classification.};

  \filldraw[black] (0.9,3.7) circle (4pt);
  \filldraw[white] (0.9,3.7) circle (2pt);
  \node[rotate=-90] at (0.5,3.7) {2023};
  \node[textNode] at (1.1,3.7) {Hochuli et al. showed that state-of-the-art deep neural networks can reach 92.8\% accuracy under cross-dataset scenarios without fine-tuning.};

  \filldraw[black] (0.9,2.6) circle (4pt);
  \filldraw[white] (0.9,2.6) circle (2pt);
  \node[rotate=-90] at (0.5,2.6) {2025};
  \node[textNode] at (1.1,2.6) {Alves et al. trained strong general teacher models that are distilled into lightweight students, reaching average accuracies of 96.6\%.};

  \filldraw[black] (0.9,1.4) circle (4pt);
  \filldraw[white] (0.9,1.4) circle (2pt);
  \node[rotate=-90] at (0.5,1.4) {2026};
  \node[textNode] at (1.1,1.4) {Santos et al. achieved 97\% accuracy using GAN-augmented target-domain data derived from 256 labeled samples, albeit with high computational cost.};

\end{tikzpicture}} 
   \caption{Timeline of works related to the parking space classification problem focusing on cross-dataset approaches.}  
   \label{fig:timeline}   
   \vspace{-5mm}
\end{figure}

Numerous strategies for parking lot classification have been explored in prior work, as comprehensively reviewed in~\cite{Almeida2022}. However, most methods are evaluated under closed-domain assumptions, with only a limited number explicitly addressing domain adaptation, particularly under scarce labeled target data. To ensure objective comparison and reproducibility, our discussion is restricted to approaches evaluated on publicly available datasets. A graphical timeline is shown in Figure~\ref{fig:timeline}.

One of the earliest works on parking lot occupancy classification was proposed by Almeida et al.~\cite{almeidaEtAl2015}, who introduced the PKLot dataset, consisting of approximately 700,000 manually annotated images captured from two parking facilities and three fixed camera viewpoints. Their approach combined hand-crafted texture descriptors with Support Vector Machine (SVM) classifiers, achieving accuracies above 99\% when extensive labeled data from the target parking lot were available for training. However, the study also highlighted limited cross-domain generalization, with performance dropping to around 90\% when models were applied to unseen parking lots without explicit domain adaptation.

Using a cross-view evaluation protocol without domain adaptation, Amato et al.~\cite{amatoEtAl2017} reported an average accuracy of 88.5\% on the CNRPark-EXT dataset, which contains approximately 160,000 labeled images acquired from nine camera viewpoints within a single parking lot. The dataset exhibits substantial visual variability due to viewpoint changes, illumination variations, shadow effects, and partial occlusions from surrounding structures.

In a related study, Nurullayev and Lee~\cite{nurullayevLee2019} proposed a deep learning framework based on dilated convolutions and pixel-skipping strategies to enlarge the effective receptive field, achieving an average accuracy of 96.4\% under a cross-view evaluation protocol without explicit domain adaptation.

Recently, Almeida et al.~\cite{Almeida2022} reported that state-of-the-art approaches reach approximately 92\% accuracy when no target-domain samples are used. Further, Hochuli et al.~\cite{hochuli2023} showed that a global model trained on PKLot and CNRPark-EXT generalizes to unseen parking lots with an average accuracy of 92.8\%, further highlighting the challenges of cross-domain generalization in parking occupancy classification.

With respect to data annotation, state-of-the-art methods typically rely on rotated rectangles or polygonal annotations to precisely delimit parking spots, which are costly to obtain. Hochuli et al.~\cite{hochuli2022} showed that simpler bounding-box annotations can achieve competitive performance by exploiting contextual information, reducing annotation effort, and reaching up to 97\% accuracy using 1{,}000 labeled samples.

Alves et al. \cite{Luza2025PKLOTDISTILLING} developed a strong general teacher models trained on thousands of labeled samples, which can be distilled into lightweight students. By distilling these students with pseudo-labeled samples collected from the target domain, using images from the first seven days of deployment, the authors achieved an average accuracy of~96.6\%. 

More recently, Santos et al. \cite{SantosPKLOTGAN} explored the use of generative adversarial networks (GANs) to synthesize target-domain samples from a limited set of labeled examples, achieving approximately 97\% accuracy using 256 annotated target samples. However, this performance depends on a complex and computationally expensive training pipeline.

As discussed so far, most state-of-the-art approaches rely on large annotated source datasets to train supervised models that are later adapted to target domains with limited labeled data. Self-taught learning (STL) addresses this limitation by learning generic representations from unlabeled data using unsupervised models, typically autoencoders, and reusing them as fixed feature extractors for task-specific classifiers. While STL has shown strong generalization capabilities in limited-data settings, such as facial expression recognition~\cite{Delazeri2025Representation}, its effectiveness for parking lot classification remains unexplored.

\begin{figure*}[htb]
    \centering
    \includegraphics[width=0.95\linewidth]{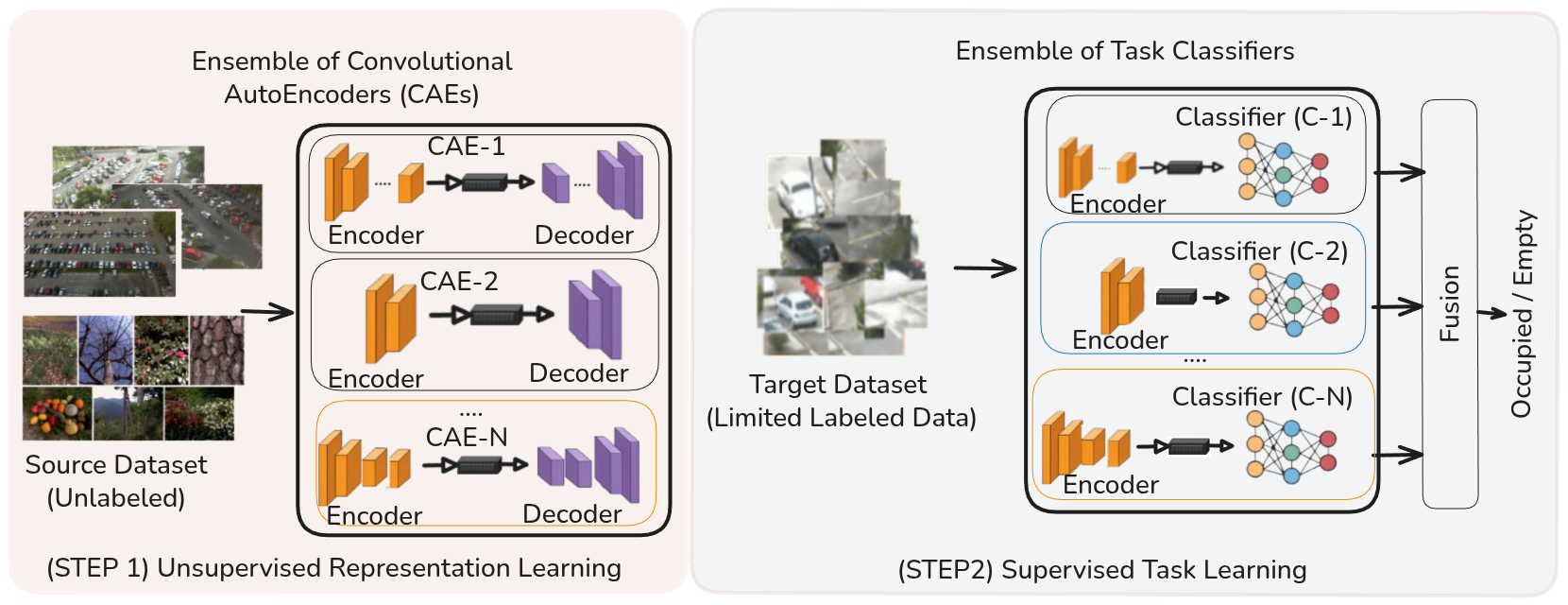}
    \caption{Proposed method: an ensemble of CAEs learns unsupervised representations from unlabeled source-domain data (Step~1), which are used to train lightweight target-domain classifiers with limited labels (Step~2). Final predictions are obtained by fusing classifier posterior probabilities.}
    \label{fig:proposed_method}
    \vspace{-4mm}
\end{figure*}

\section{Proposed Method}\label{sec:proposed_method}

To reduce data acquisition and labeling costs while improving generalization under cross-dataset parking scenarios, this work proposes an ensemble-based self-taught learning (STL) framework for parking lot applications. The key contribution is the use of an ensemble of heterogeneous convolutional autoencoders to learn transferable visual representations in a fully unsupervised manner. Architectural diversity across autoencoders enables the learning of complementary latent features, mitigating architectural bias and improving robustness to domain shifts, as evidenced in Delazeri et al.~\cite{Delazeri2025Representation}. These representations are subsequently reused to train lightweight classifiers with limited labeled target data, enabling scalable and efficient deployment.

Figure~\ref{fig:proposed_method} illustrates the proposed framework. In the representation learning stage (Step~1), the ensemble of convolutional autoencoders is trained independently on unlabeled source datasets. In the task learning stage (Step~2), encoder weights are frozen and reused as fixed feature extractors, while lightweight classifiers are trained on top of the latent representations using a limited number of labeled target samples. During inference, predictions from all encoder-classifier pairs are combined through an ensemble fusion rule that aggregates posterior probabilities to produce the final decision.

To the best of our knowledge, this approach has not yet been explored in the context of parking lot analysis. Additionally, we analyze the impact of the source domain, varying levels of similarity to the target domain, the effect of reducing annotated target samples, and the generalization capability of the proposed method.

\subsection{Representation Learning}\label{sec:repres_learning}
In the representation learning stage, multiple convolutional autoencoders (CAEs) are independently trained to capture diverse and complementary visual representations from unlabeled data. Each CAE follows an encoder–decoder design with progressive spatial downsampling in the encoder and a symmetric decoder for reconstruction.

We employ a stochastic generator that instantiates convolutional architectures (encoder) with heterogeneous depth, downsampling strategies, and latent dimensionality. The generation procedure is detailed in Algorithm~\ref{alg:stochastic_cae}, while the architectural hyperparameters follow the guidelines proposed in Delazeri et al.~\cite{Delazeri2025Representation}. The number of generated CAEs and the corresponding training protocol are described in Section~\ref{sec:experiments}.

\begin{algorithm}[!htbp]
\caption{Stochastic Convolutional Autoencoder Generator}
\label{alg:stochastic_cae}
\begin{algorithmic}[1]
\State Sample number of encoder blocks $L \sim \mathcal{U}\{2,5\}$
\State Initialize pooling counter $p \leftarrow 0$
\For{$i = 1$ to $L$}
    \State Sample filters $f_i \in \{8,16,32,64,128\}$
    \State Add $\text{Conv2D}(3\times3, f_i)$ + ReLU
    \If{$\mathcal{U}(0,1) < .5$ \textbf{and} $p<4$}
        \State Add MaxPooling2D, $p \leftarrow p+1$
    \EndIf
\EndFor
\State Sample latent dimension $d \sim \mathcal{U}[128,2048]$
\State Project encoder output to latent vector $\mathbf{z}\in\mathbb{R}^d$
\State Construct decoder symmetrically to encoder
\For{each encoder block in reverse order}
    \State Add Conv2DTransp (stride=2 if pooled else 1)
\EndFor
\State Add final $\text{Conv2D}(3\times3)$ with sigmoid activation
\end{algorithmic}
\end{algorithm}

For each instance, the number of convolutional blocks is sampled from the interval $\{2,5\}$. Each block consists of a $3\times3$ \textit{Conv2D} layer with ReLU activation, where the number of filters is randomly selected from~$\{8,16,32,\allowbreak64,\allowbreak128\}$. After each convolutional layer, a \textit{MaxPooling2D} operation is inserted with probability $0.5$. To prevent excessive spatial collapse for $64\times64$ inputs, the total number of pooling operations is explicitly constrained to a maximum of four. The convolutional sequence output is mapped to a latent representation of dimension $d$, where $d$ is randomly sampled from the range $\{128,2048\}$.

The decoder is constructed symmetrically to the encoder. The latent vector is first expanded via a fully connected layer and reshaped to match the final encoder feature-map dimensions. A sequence of \textit{Conv2DTranspose} layers then progressively reconstructs the input, using stride $2\times2$ for stages corresponding to encoder blocks that applied max pooling, and stride $1\times1$ otherwise. The final reconstruction layer employs a $3\times3$ convolution with sigmoid activation to match normalized pixel intensities.

\subsection{Task Learning}\label{sec:task_learning}

In the task learning phase, the encoder learned during representation learning (Section~\ref{sec:repres_learning}) is reused as a fixed convolutional feature extractor, with all parameters frozen to preserve the learned representations.

The classifier is constructed by stacking a lightweight fully connected head on top of the frozen encoder, as illustrated in the Figure \ref{fig:proposed_method} (Step 2). A dropout layer with a rate of $0.2$ is first applied to the latent features to improve regularization. This is followed by two fully connected layers with 256 and 128 neurons, respectively, both using ReLU activation function. The final layer has two output units followed by Softmax, producing the posterior class probabilities.

\subsection{Training and Inference}\label{sec:train_and_infer}

Each convolutional autoencoder (CAE) in the ensemble is trained independently during the unsupervised representation learning stage. Given an unlabeled source dataset $\mathcal{D}_s = \{\mathbf{x}_i\}_{i=1}^{N_s}$, each CAE $f_{\theta_k}$, with $k=1,\dots,K$, is optimized by minimizing the mean squared reconstruction error:
\[
\mathcal{L}_{\mathrm{rec}}^{(k)} =
\frac{1}{|\mathcal{D}_s|}
\sum_{\mathbf{x}\in\mathcal{D}_s}
\left\| \mathbf{x} - f_{\theta_k}(\mathbf{x}) \right\|_2^2,
\]
where $f_{\theta_k}(\mathbf{x}) = D_{\theta_k}(E_{\theta_k}(\mathbf{x}))$, and $E_{\theta_k}$ and $D_{\theta_k}$ denote the encoder and decoder of the $k$-th CAE, respectively. Each CAE is trained using the Adam optimizer, without parameter sharing or joint optimization, promoting the learning of different latent representations across the ensemble.

Given a labeled target dataset $\mathcal{D}_t = \{(\mathbf{x}_j, y_j)\}_{j=1}^{N_t}$, a lightweight classifier $g_{\phi_k}$ is trained on top of the extracted latent features $\mathbf{z}_j^{(k)} = E_{\theta_k}(\mathbf{x}_j)$. The classifier head is optimized using the Adam optimizer by minimizing the categorical cross-entropy loss.

At inference time, the ensemble produces a set of predictions $\{\hat{y}^{(k)}\}_{k=1}^{K}$, one for each CAE--classifier pair.
These predictions are aggregated using a majority voting rule
\[
\hat{y} =
\arg\max_{c \in \{1,\dots,C\}}
\sum_{k=1}^{K}
\mathbb{I}\!\left(\hat{y}^{(k)} = c\right),
\]
yielding the final class decision.

\subsection{Datasets}\label{sec:datasets}

In this work, we employ two widely used benchmark datasets for vision-based parking occupancy classification: (i) PKLot~\cite{almeidaEtAl2015} and (ii) CNRPark-EXT~\cite{amatoEtAl2017}. Both datasets reflect real-world deployment conditions by incorporating multiple parking facilities and substantial environmental variability.

\begin{figure}[!htbp]
  \centering    
  \subfloat[UFPR04]{\includegraphics[height=1.5cm,width=2.85cm,trim={00px 0px 0px 0px}, clip]{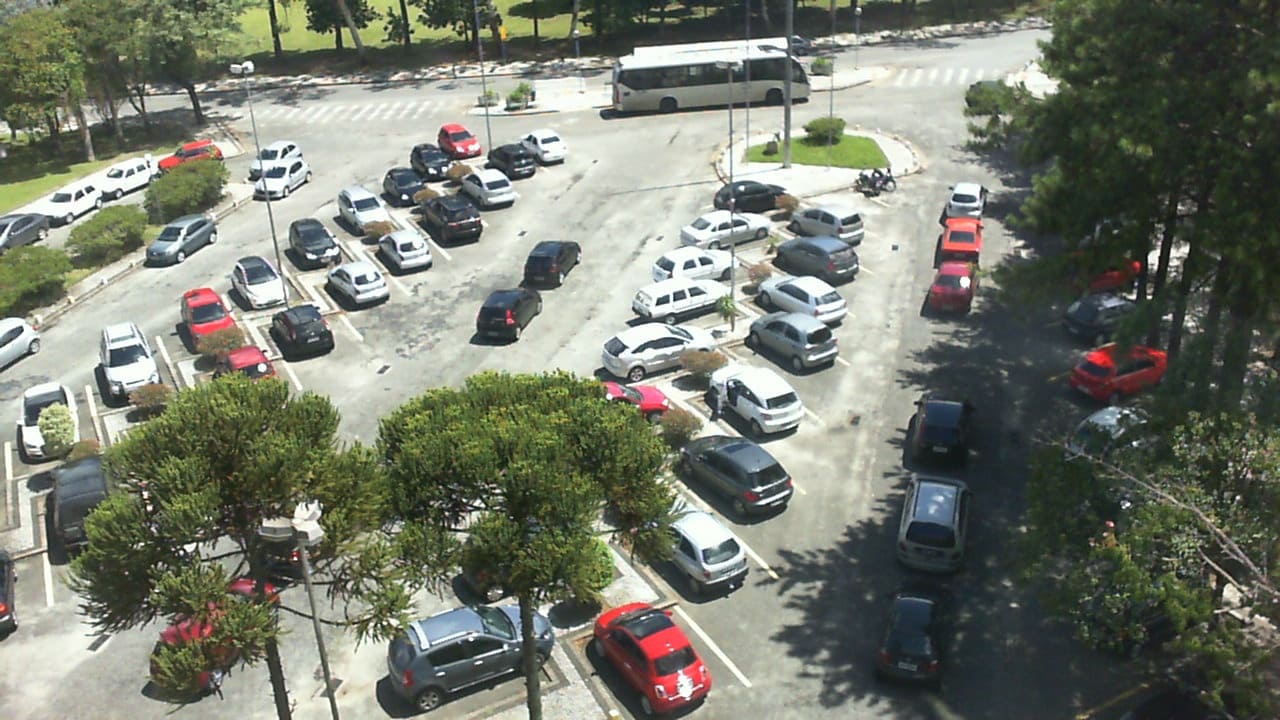}}\hfill
  \subfloat[UFPR05]{\includegraphics[height=1.5cm,width=2.85cm,trim={00px 0px 0px 0px}, clip]{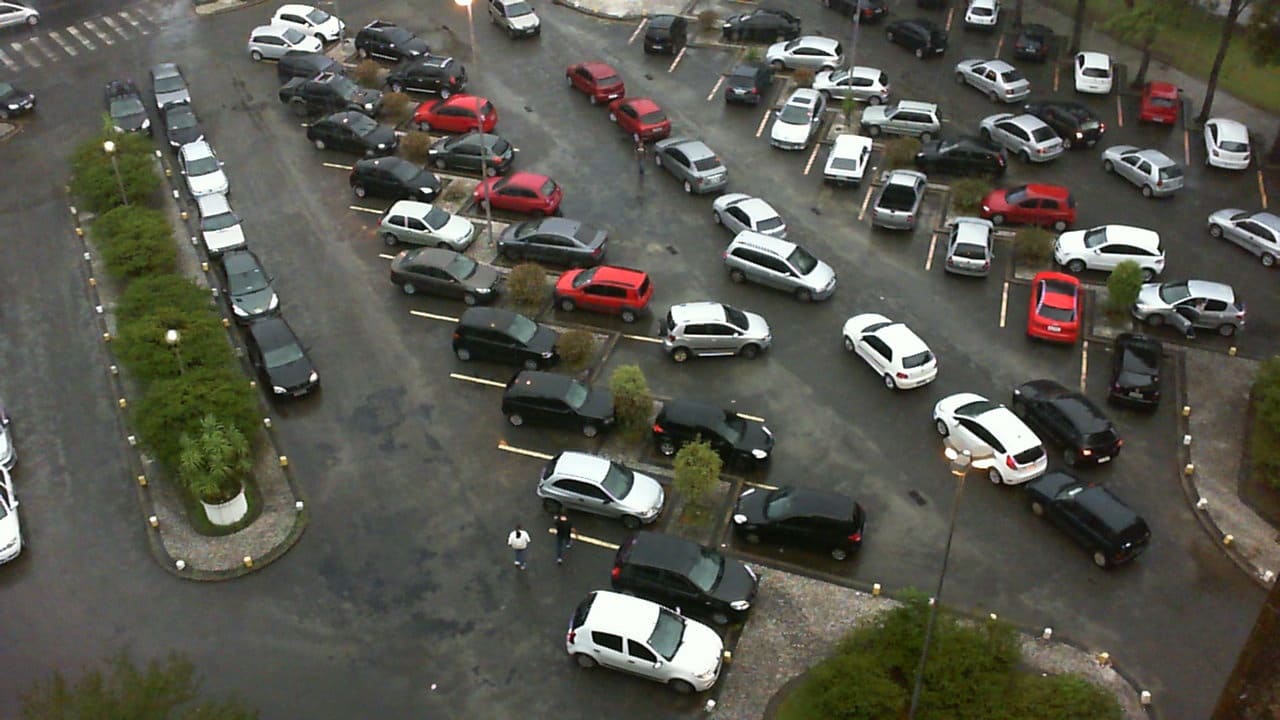}}\hfill
  \subfloat[PUCPR]{\includegraphics[height=1.5cm,width=2.85cm, trim={00px 0px 0px 0px}, clip]{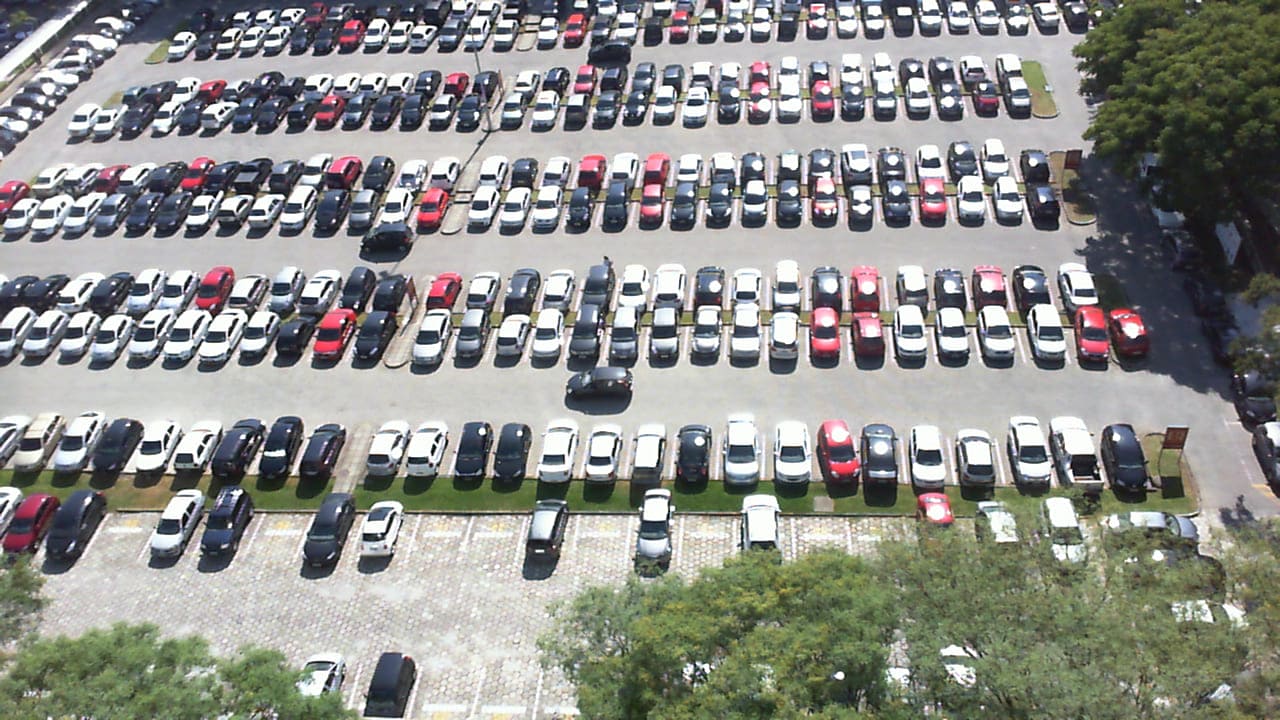}}\quad
  \caption{The three deployment scenarios from PKLot.}
  \label{fig:overview_pklot} 
\end{figure}

\textbf{PKLot.} The PKLot dataset \cite{almeidaEtAl2015} was acquired over a period of roughly three months using a fixed temporal sampling rate of five minutes between consecutive captures. It contains 12,417 images and approximately 700,000 manually labeled parking spaces, organized into three independent parking lot scenarios: UFPR04, UFPR05, and PUCPR. Image examples of the dataset are shown in Figure~\ref{fig:overview_pklot}.

\begin{figure}[!ht]
 
  \centering    
  \subfloat[Camera 1]{\includegraphics[height=1.5cm,width=2.85cm,trim={0px 0px 0px 0px}, clip]{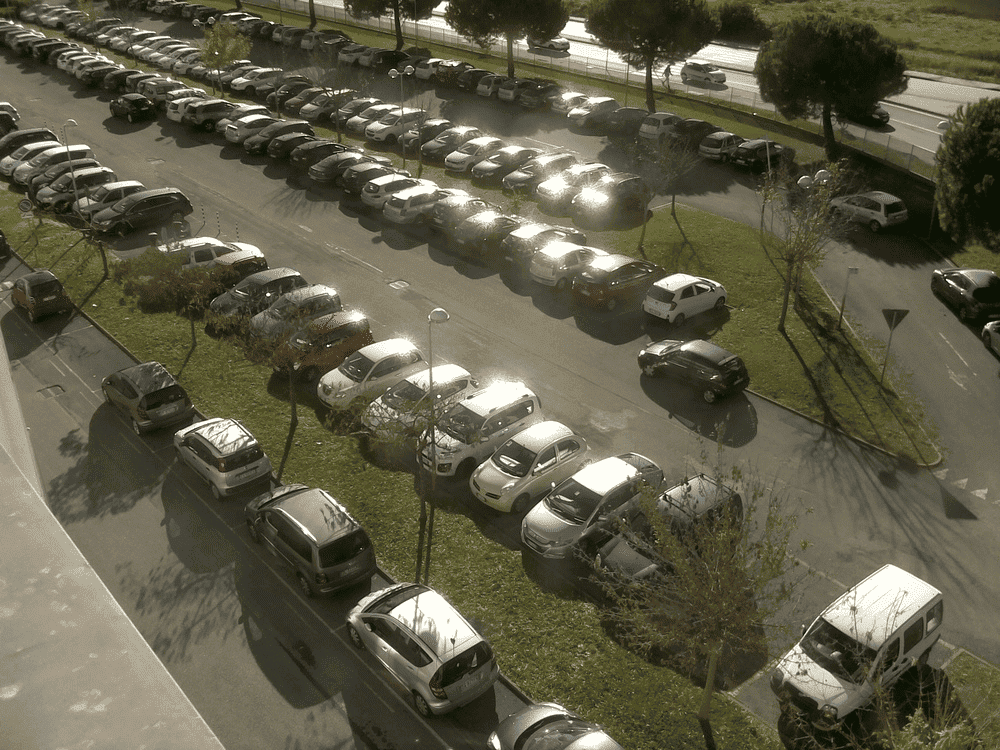}} \hfill
  \subfloat[Camera 4]{\includegraphics[height=1.5cm,width=2.85cm, trim={0px 0px 0px 0px}, clip]{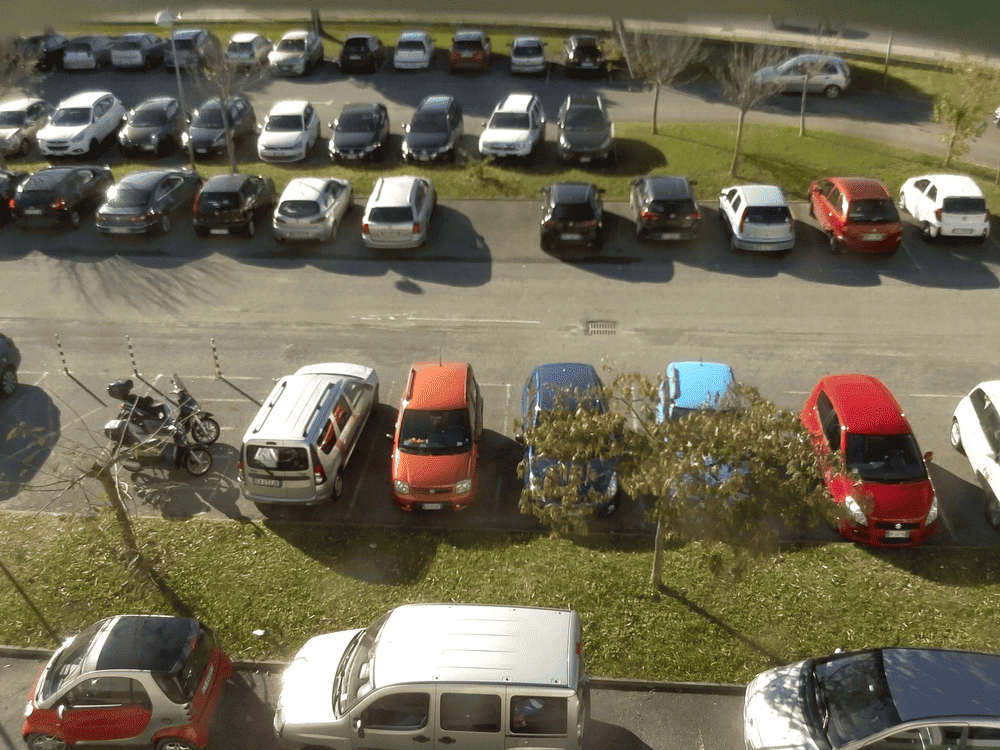}} \hfill
  \subfloat[Camera 9]{\includegraphics[height=1.5cm,width=2.85cm,trim={0px 0px 0px 70px}, clip]{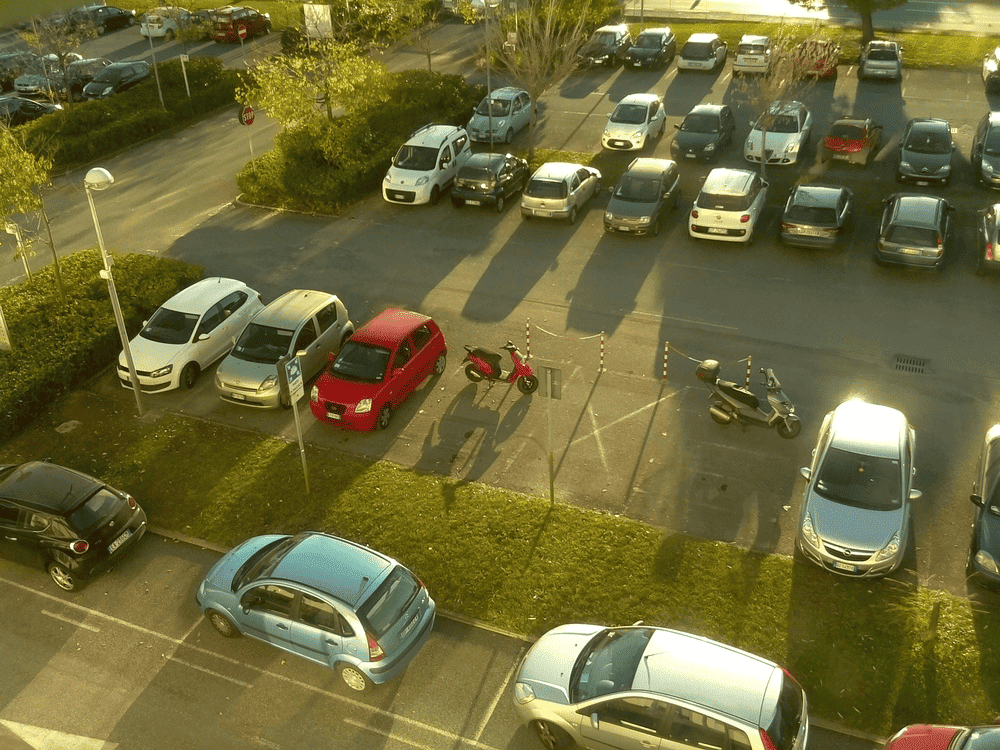}} \hfill
    
  \caption{Three out of nine scenarios from the CNRPark-EXT.}
  \label{fig:overview_cnr}  

\end{figure}

\textbf{CNRPark-EXT.} Introduced by Amato et al. \cite{amatoEtAl2017}, the CNRPark-EXT dataset includes approximately 160,000 labeled parking spaces captured from nine static camera viewpoints observing a single parking lot. The dataset poses several challenges, such as severe illumination artifacts, environmental noise from raindrops on the camera lens, and frequent partial occlusions from trees and lamp posts. Image examples are presented in Figure~\ref{fig:overview_cnr}.

Table \ref{tab:parking_spots_per_dataset} briefly summarizes the properties of the datasets. For a thorough description of the datasets and their applications, refer to  \cite{Almeida2022, almeidaEtAl2015, amatoEtAl2017}.

\begin{table}[!h]
\setlength{\tabcolsep}{4pt}
\centering
\caption{Summary parking lots datasets used in this work}
\label{tab:parking_spots_per_dataset}
\begin{tabular}{ccccc}
\hline
\textbf{Dataset}  & \textbf{Spots} & \textbf{Days} & \textbf{Camera Angles} & \textbf{Park. Lots}\\ 
\hline
PKLot  & 695,851 & 99 & 3 & 2\\ 
CNRPark-EXT & 157,549 & 23 & 9 & 1\\
\hline
\vspace{-2mm}
\end{tabular}

\end{table}

\section{Experiments}\label{sec:experiments}
\label{sec:experiments}
In this section, we evaluate the proposed approach by adopting a widely used cross-dataset evaluation protocol~\cite{amatoEtAl2017, Almeida2022, hochuli2023, hochuli2022, nurullayevLee2019, SantosPKLOTGAN} to ensure fair comparison with state-of-the-art methods.
Using the datasets described in Section~\ref{sec:datasets}, the cross-dataset protocol enforces a strict separation between datasets used for unsupervised representation learning and those used for target-domain task-specific adaptation. This design prevents data bias by ensuring that samples from scenarios observed during representation learning are not reused during target-domain optimization. Each target camera is treated as an independent deployment scenario, yielding 12 target scenarios under heterogeneous conditions. All models are trained following the procedures and hyperparameters defined in Section~\ref{sec:train_and_infer}.

\subsection{Unsupervised Representations Learning (STEP 1)}\label{sec:cae_training}

For unsupervised representation learning, ten convolutional autoencoders (\(K=10\)) were generated using the stochastic architecture generator described in Section~\ref{sec:repres_learning}. Each autoencoder ($f_{\theta_k}^{\ast}$) was trained independently for 50 epochs, and the model achieving the lowest validation MSE was selected. Training was conducted using \(1{,}000\) randomly sampled images and \(64\) validation images per representation dataset, ensuring balanced exposure to representation-domain variability. As representation learning is performed in an unsupervised manner, this subset size is sufficient to capture dominant structural patterns without introducing dataset-specific bias.

\begin{figure}[!htbp]
    \centering
    \includegraphics[width=0.9\linewidth]{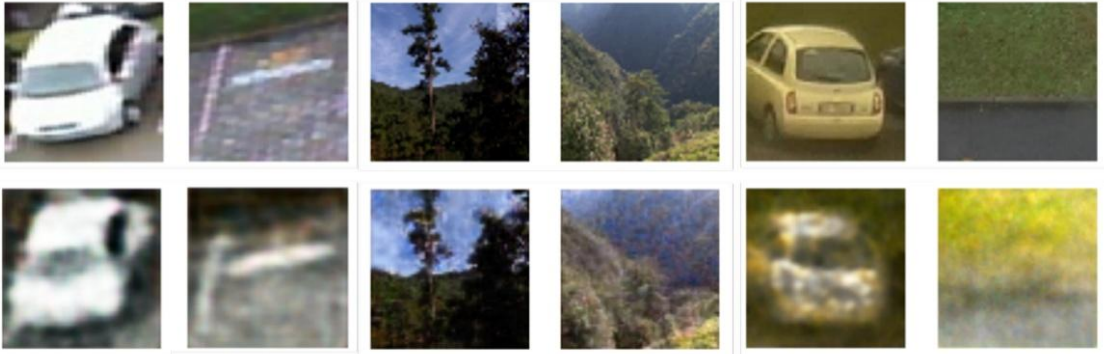}
    \caption{Qualitative comparison between real samples (top) and reconstructed ones (bottom).} 
    \label{fig:rec_results}
\end{figure}

Figure~\ref{fig:rec_results} illustrates reconstruction results on representation-domain scenarios. Although some geometric distortions are present, the primary goal is to expose the representation learning process to diverse shapes and textures~\cite{baker2018shape}. This diversity is particularly relevant, as the downstream task focuses on occupancy-related visual cues rather than object identity (e.g., vehicle models).

\subsection{Supervised Task Learning (STEP 2)}\label{sec:cae_training}
In the supervised task learning stage, each selected encoder ($E_{\theta_k}$) was reused as a fixed feature extractor and paired with an independent classifier head $g_{\phi_k}$, as discussed in Section \ref{sec:task_learning}. All encoder parameters were frozen. Each classifier ($g_{\phi_k}^{\ast}$) was trained independently for 10 epochs, and the model achieving the lowest validation loss was selected.

\begin{figure}[!htbp]  
  \centering    

  \includegraphics[width=1.6cm,height=1.1cm]{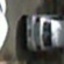}
  \includegraphics[width=1.6cm,height=1.1cm]{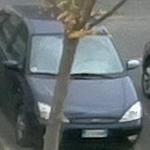}
  \includegraphics[width=1.6cm,height=1.1cm]{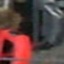}
  \includegraphics[width=1.6cm,height=1.1cm]{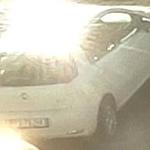}
  \includegraphics[width=1.6cm,height=1.1cm]{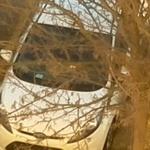}
  \caption{Examples individual parking spots across the camera views in the PKLot and CNRPark-EXT datasets.}
  
  \label{fig:cnr_challenges}  

\end{figure}

Figure~\ref{fig:cnr_challenges} illustrates representative parking spaces highlighting the domain shifts considered in our evaluation.

To address RQ1, which investigates the impact of the number of annotated target samples ($R$) on effective task adaptation, we conducted experiments using $R \in \{64,128,256,512,1024\}$ labeled samples from the target domain. These samples were used exclusively in the supervised task learning stage. Samples were collected chronologically, starting from the first day of acquisition for each target camera, and all samples from these days were excluded from the test set to prevent data leakage. The values of $R$ are selected based on prior studies~\cite{hochuli2022,SantosPKLOTGAN}, which show that competitive parking occupancy classification performance can be achieved with approximately 1{,}000 annotated target samples, with marginal gains beyond this point. 

Table~\ref{tab:batch_vs_base} reports classification accuracy across target scenarios. Accuracy consistently improves with additional annotated samples, with several scenarios achieving strong performance under limited supervision (64–128 samples). In most cases, gains saturate beyond 256 samples, corroborating prior observations~\cite{SantosPKLOTGAN}. More challenging camera views show greater sensitivity to annotation size, while overall results remain stable, underscoring the robustness and transferability of the proposed approach across heterogeneous target conditions.

\begin{table}[!htbp]
\centering
\caption{Classification accuracy (\%) achieved by task-specific classifiers across different target scenarios as a function of the number of annotated target samples}
\label{tab:batch_vs_base}
\resizebox{0.9\linewidth}{!}{
\begin{tabular}{ccccccc}
\hline
\multirow{2}{*}{\textbf{\begin{tabular}[c]{@{}c@{}}Representation \\ Domain\end{tabular}}} & \multirow{2}{*}{\textbf{\begin{tabular}[c]{@{}c@{}}Target \\ Domain\end{tabular}}} & \multicolumn{5}{c}{\textbf{Annotated Data}} \\
 &  & \textbf{64} & \textbf{128} & \textbf{256} & \textbf{512} & \textbf{1024} \\ \hline
\multirow{3}{*}{\textbf{CNR}} & PUC & 95.1 & 97.4 & 98.2 & 98.4 & 98.7 \\
 & UFPR04 & 94.0 & 96.3 & 96.5 & 97.0 & 97.4 \\
 & UFPR05 & 97.0 & 97.6 & 98.1 & 98.4 & 98.3 \\ \hline
\multirow{9}{*}{\textbf{PKLOT}} & CNR-1 & 90.7 & 92.6 & 93.7 & 94.3 & 95.8 \\
 & CNR-2 & 97.3 & 97.1 & 97.4 & 97.6 & 98.4 \\
 & CNR-3 & 95.8 & 94.7 & 96.3 & 97.0 & 97.7 \\
 & CNR-4 & 94.7 & 94.5 & 95.1 & 95.0 & 95.4 \\
 & CNR-5 & 86.2 & 89.7 & 92.3 & 93.5 & 94.7 \\
 & CNR-6 & 92.2 & 92.5 & 93.7 & 93.4 & 95.1 \\
 & CNR-7 & 92.3 & 91.8 & 91.1 & 92.0 & 92.9 \\
 & CNR-8 & 92.9 & 95.0 & 95.3 & 96.0 & 96.6 \\
 & CNR-9 & 91.2 & 91.0 & 92.5 & 94.3 & 95.8 \\ \hline
\multicolumn{2}{c}{\textbf{Average}} & \textbf{93.3} & \textbf{94.2} & \textbf{95.0} & \textbf{95.6} & \textbf{96.4} \\ \hline
\end{tabular}}
\end{table}

An important aspect is the ability to generalize to unseen environments (RQ2). Accordingly, task-specific classifiers trained on the target domain using 1{,}024 annotated samples, the amount yielding the best performance, are evaluated across the remaining scenarios. The results are presented in Table~\ref{tab:cross_domain_voting_percent}.

\begin{table*}[!htb]
\centering
\setlength{\tabcolsep}{3pt} %
\caption{Cross-scenario performance (accuracy \%). The first column denotes the classifier domain and the remaining columns correspond to the target cross-domains.}

\label{tab:cross_domain_voting_percent}
\resizebox{0.75\linewidth}{!}{
\begin{tabular}{ccccccccccccc}
\toprule
\multirow{2}{*}{\textbf{\begin{tabular}[c]{@{}c@{}}Task-Specific \\ Classifier\end{tabular}}} & \multicolumn{12}{c}{\textbf{Cross Domains (without weight optimization)}} \\
\cmidrule(lr){2-13}
 & CNR-1 & CNR-2 & CNR-3 & CNR-4 & CNR-5 & CNR-6 & CNR-7 & CNR-8 & CNR-9 & PUC & UFPR04 & UFPR05 \\
\midrule
CNR-1  & 95.8 & 98.1 & 96.5 & 96.1 & 92.0 & 95.7 & 90.9 & 95.7 & 93.7 & 89.2 & 76.9 & 76.6 \\
CNR-2  & 90.5 & 98.4 & 95.2 & 96.3 & 91.4 & 95.8 & 90.3 & 96.2 & 93.8 & 91.2 & 85.4 & 84.1 \\
CNR-3  & 85.9 & 98.0 & 97.7 & 96.0 & 89.0 & 95.1 & 87.3 & 95.9 & 91.5 & 92.3 & 83.0 & 87.4 \\
CNR-4  & 86.3 & 97.2 & 94.1 & 95.4 & 90.8 & 95.3 & 87.7 & 96.5 & 92.7 & 93.7 & 88.3 & 87.4 \\
CNR-5  & 88.5 & 98.3 & 96.3 & 97.2 & 94.7 & 96.1 & 90.7 & 96.8 & 93.8 & 96.6 & 88.4 & 85.4 \\
CNR-6  & 83.1 & 96.5 & 93.0 & 94.5 & 90.1 & 95.1 & 86.0 & 95.5 & 90.1 & 93.8 & 85.5 & 84.2 \\
CNR-7  & 91.1 & 98.1 & 97.7 & 97.6 & 95.4 & 97.2 & 92.9 & 97.3 & 95.4 & 93.3 & 90.1 & 79.4 \\
CNR-8  & 83.0 & 93.9 & 94.6 & 93.7 & 89.5 & 95.3 & 87.3 & 96.6 & 92.0 & 94.8 & 84.7 & 85.7 \\
CNR-9  & 91.1 & 98.0 & 95.4 & 97.3 & 94.0 & 97.0 & 91.0 & 97.6 & 95.8 & 91.8 & 90.1 & 88.2 \\
PUC      & 91.4 & 95.0 & 92.0 & 95.4 & 92.4 & 94.4 & 90.2 & 93.4 & 91.4 & 98.7 & 91.5 & 81.3 \\
UFPR04   & 89.0 & 94.2 & 92.8 & 92.6 & 88.5 & 93.6 & 90.6 & 91.8 & 88.5 & 93.2 & 97.4 & 84.5 \\
UFPR05   & 71.7 & 83.0 & 85.1 & 84.9 & 74.9 & 87.6 & 76.9 & 85.8 & 77.4 & 82.6 & 87.4 & 98.3 \\
\bottomrule
\end{tabular}}
\vspace{-2mm}
\end{table*}

The proposed ensemble-based self-taught learning framework demonstrates feasible generalization to domain shifts, without additional weight optimization. High accuracy is consistently achieved across target scenarios, particularly among cameras within the same dataset, indicating effective transfer of learned representations. Performance degradation is more evident when transferring between datasets with markedly different acquisition conditions, such as CNRPark-EXT and PKLot, reflecting severe shifts. Nevertheless, competitive accuracy is maintained in most cross-domain settings, confirming the robustness and transferability of the learned representations and validating the proposed approach in realistic deployment scenarios (RQ2).

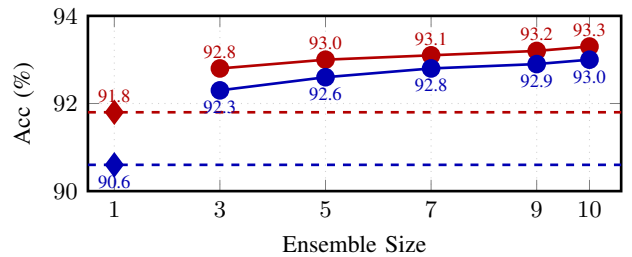
\begin{figure}[!htbp]
\centering
\begin{tikzpicture}
\begin{axis}[
    width=\columnwidth,
    height=3.9cm,
    xmin=0.5, xmax=10.6,
    ymin=90, ymax=94,
    xtick={1,3,5,7,9,10},
    ytick={90,92,94},
    xlabel={Ensemble Size},
    ylabel={Acc (\%)},
    grid=both,
    grid style={dotted, gray!35},
    tick label style={font=\small},
    label style={font=\small},
    line width=1pt,
]

\addplot[
    dashed,
    color=red!70!black,
] coordinates {(0.5,91.8) (10.6,91.8)};

\addplot[
    dashed,
    color=blue!70!black,
] coordinates {(0.5,90.6) (10.6,90.6)};

\addplot[
    color=red!70!black,
    mark=*,
    mark size=3pt,
] coordinates {
    (3,92.8)
    (5,93.0)
    (7,93.1)
    (9,93.2)
    (10,93.3)
};

\addplot[
    color=blue!70!black,
    mark=*,
    mark size=3pt,
] coordinates {
    (3,92.3)
    (5,92.6)
    (7,92.8)
    (9,92.9)
    (10,93.0)
};

\addplot[
    only marks,
    mark=diamond*,
    mark size=4pt,
    color=red!70!black,
] coordinates {(1,91.8)};

\addplot[
    only marks,
    mark=diamond*,
    mark size=4pt,
    color=blue!70!black,
] coordinates {(1,90.6)};

\node[font=\scriptsize, color=red!70!black, above] at (axis cs:1,91.8) {91.8};
\node[font=\scriptsize, color=blue!70!black, below] at (axis cs:1,90.6) {90.6};

\node[font=\scriptsize, color=red!70!black, above] at (axis cs:3,92.8) {92.8};
\node[font=\scriptsize, color=red!70!black, above] at (axis cs:5,93.0) {93.0};
\node[font=\scriptsize, color=red!70!black, above] at (axis cs:7,93.1) {93.1};
\node[font=\scriptsize, color=red!70!black, above] at (axis cs:9,93.2) {93.2};
\node[font=\scriptsize, color=red!70!black, above] at (axis cs:10,93.3) {93.3};

\node[font=\scriptsize, color=blue!70!black, below] at (axis cs:3,92.3) {92.3};
\node[font=\scriptsize, color=blue!70!black, below] at (axis cs:5,92.6) {92.6};
\node[font=\scriptsize, color=blue!70!black, below] at (axis cs:7,92.8) {92.8};
\node[font=\scriptsize, color=blue!70!black, below] at (axis cs:9,92.9) {92.9};
\node[font=\scriptsize, color=blue!70!black, below] at (axis cs:10,93.0) {93.0};

\end{axis}
\end{tikzpicture}
\caption{Ensemble size vs. accuracy (\%) averaged over all dataset scenarios (red: CNR, blue: PKLot). Solid lines with circular markers ($\bullet$) denote ensemble performance; dashed lines (--) denote single-model baselines ($\diamond$).}
\label{fig:ensemble-ablation}
\vspace{-3mm}
\end{figure}

One may question the choice of using 10 autoencoders. Figure \ref{fig:ensemble-ablation} shows that cross domain generalization performance, averaged across all dataset scenarios, increases with ensemble size but quickly saturates, reaching a plateau around 10 models. Beyond this point, improvements are negligible, indicating that 10 autoencoders effectively approximate the upper performance bound while maintaining a favorable cost benefit trade off.

\subsection{State-of-Art Comparison}

Table~\ref{tab:sota_comparison} presents a comparison between the proposed framework and state-of-the-art methods (RQ3), focusing on cross-dataset evaluation, reliance on target-domain supervision, and annotation requirements. As can be observed, even with only 64 annotated samples, the proposed method maintains an average accuracy of about 93\%, outperforming several approaches.

\begin{table}[!htpb]
\centering
\setlength{\tabcolsep}{1.8pt}
\caption{State-of-the-Art Comparison. In $^1$ the target samples used are unlabeled.}
\label{tab:sota_comparison}
\resizebox{\linewidth}{!}{%
\begin{tabular}{ccccc}
\toprule
\textbf{Author} & \textbf{Approach} & \textbf{\begin{tabular}[c]{@{}c@{}}Limited Target \\ Annot. Data\end{tabular}} & \textbf{\begin{tabular}[c]{@{}c@{}}Target\\ Training\end{tabular}} & \textbf{Accuracy} \\
Almeida et al.\cite{almeidaEtAl2015} & LBP + SVM & - & No & $\sim$87\% \\
Amato et al.\cite{amatoEtAl2017} & Custom CNN & - & No & $\sim$88\% \\
Nurullayev et al.\cite{nurullayevLee2019} & Custom CNN & - & No & $\sim$96\% \\
Almeida et al.\cite{Almeida2022} & Survey & - & - & $\sim$92\% \\
Hochuli et al.\cite{hochuli2023} & MobileNetV3  & - & No & $\sim$92\% \\
Alves et al.\cite{Luza2025PKLOTDISTILLING} & \begin{tabular}[c]{@{}c@{}}MobileNetV3 \\ and Custom CNN\end{tabular}  & - & Yes$^1$ & $\sim$97\% \\
Santos et al. \cite{SantosPKLOTGAN} & \begin{tabular}[c]{@{}c@{}}GANs \\ and MobileNetV3\end{tabular} & 1024 samples& Yes & $\sim$98\% \\ \hline
\textbf{Ours} & \multicolumn{1}{c}{\textbf{\textbf{\begin{tabular}[c]{@{}l@{}}Ensemble \\ CAE-ANN\end{tabular}}}} & \textbf{\begin{tabular}[c]{@{}c@{}}64 samples\\ 1024 samples\end{tabular}} & \textbf{Yes} & \textbf{\begin{tabular}[c]{@{}c@{}}$\sim$93\%\\ $\sim$96\%\end{tabular}} \\ \bottomrule
\end{tabular}%
}
\vspace{-5mm}
\end{table}

Overall, the results show that the proposed ensemble-based CAE framework achieves competitive performance with substantially reduced annotation requirements. While most state-of-the-art methods rely on large annotated datasets to learn visual representations, our approach learns representations in an unsupervised manner and requires only limited supervision for task adaptation. With 1{,}024 annotated target samples, it achieves an average accuracy of approximately 96\%, within two percentage points of the GAN-based method of Santos et al.~\cite{SantosPKLOTGAN}, while requiring only the training of the classifier head during deployment.

Compared to Alves et al. \cite{Luza2025PKLOTDISTILLING}, which proposed a recent approach with competitive results, our method does not require several days (e.g., 7 days) of target data collection, and does not require the training/fine-tunning of the entire classifier during deployment (only the classification head). However, it has the drawback that, although it requires only a small number of samples, these must be manually labeled, whereas Alves et al. rely on pseudo-labels from the target parking lot.

\section{Conclusion}\label{sec:conclusion}
This work introduced an ensemble-based self-taught learning framework for parking spot occupancy classification that combines unsupervised representation learning with efficient task adaptation. By training heterogeneous convolutional autoencoders on unlabeled data and reusing their encoders as fixed feature extractors, the proposed approach reduces annotation requirements while promoting robustness under domain shifts.

Experiments on the PKLot and CNRPark-EXT benchmarks under cross-dataset protocols demonstrate competitive performance across twelve target scenarios, even with limited supervision. High accuracy is achieved with as few as 64 annotated samples, and performance saturates with modest annotation effort. Cross-scenario evaluations further confirm strong generalization without additional weight optimization. Overall, the proposed framework offers an effective and scalable alternative to fully supervised methods for real-world parking monitoring applications.

\section*{Acknowledgment}

The authors acknowledge the financial support of the Coordenação de Aperfeiçoamento de Pessoal de Nível Superior (CAPES), Financiadora de Estudos e Projetos (FINEP), and the Fundação Araucária, in partnership with the Secretaria da Ciência, Tecnologia e Ensino Superior do Estado do Paraná (SETI-PR), under Grant No. 653/2025.

\balance
\bibliographystyle{IEEEtran}
\bibliography{biblio}

\end{document}